\documentclass[runningheads]{llncs}
\usepackage[T1]{fontenc}
\usepackage{graphicx}
\usepackage{amsfonts}
\usepackage{amsmath}
\usepackage{booktabs}
\usepackage{hyperref}
\usepackage{color}
\usepackage{algorithm}
\usepackage{algpseudocode}
\usepackage{longtable}
\usepackage[flushleft]{threeparttable}

\hypersetup{
    colorlinks,
    linkcolor={blue},
    citecolor={blue},
    urlcolor={blue}
}

\usepackage{cleveref}
\begin{document}
\title{3D Shape-to-Image Brownian Bridge Diffusion for Brain MRI Synthesis from Cortical Surfaces}
\titlerunning{Cor2Vox}
%

\author{
Fabian Bongratz\inst{1,2}*
Yitong Li\inst{1,2}*,
Sama Elbaroudy\inst{1},
Christian Wachinger\inst{1,2}}

\institute{Lab for Artificial Intelligence in Medical Imaging, \\ Technical University of Munich (TUM), Germany \and
Munich Center for Machine Learning (MCML), Germany} 

%
\authorrunning{F. Bongratz, Y. Li, S. Elbaroudy, C. Wachinger}
%
%
\maketitle              

\def\thefootnote{\normalsize*}\footnotetext{The authors contributed equally to this paper. \\ Email: \texttt{\{fabi.bongratz, yi\_tong.li\}@tum.de}}

\begin{abstract}
Despite recent advances in medical image generation, existing methods struggle to produce anatomically plausible 3D structures. In synthetic brain magnetic resonance images (MRIs), characteristic fissures are often missing, and reconstructed cortical surfaces appear scattered rather than densely convoluted. 
To address this issue, we introduce Cor2Vox, the first diffusion model-based method that translates continuous cortical shape priors to synthetic brain MRIs. 
To achieve this, we leverage a Brownian bridge process which allows for direct structured mapping between shape contours and medical images. Specifically, we adapt the concept of the Brownian bridge diffusion model to 3D and extend it to embrace various complementary shape representations.
Our experiments demonstrate significant improvements in the geometric accuracy of reconstructed structures compared to previous voxel-based approaches. Moreover, Cor2Vox excels in image quality and diversity, yielding high variation in non-target structures like the skull. Finally, we highlight the capability of our approach to simulate cortical atrophy at the sub-voxel level.
Our code is available at~\url{https://github.com/ai-med/Cor2Vox}.

\end{abstract}

\section{Introduction}
\begin{figure}[t]
    \centering
    \includegraphics[width=0.95\linewidth]{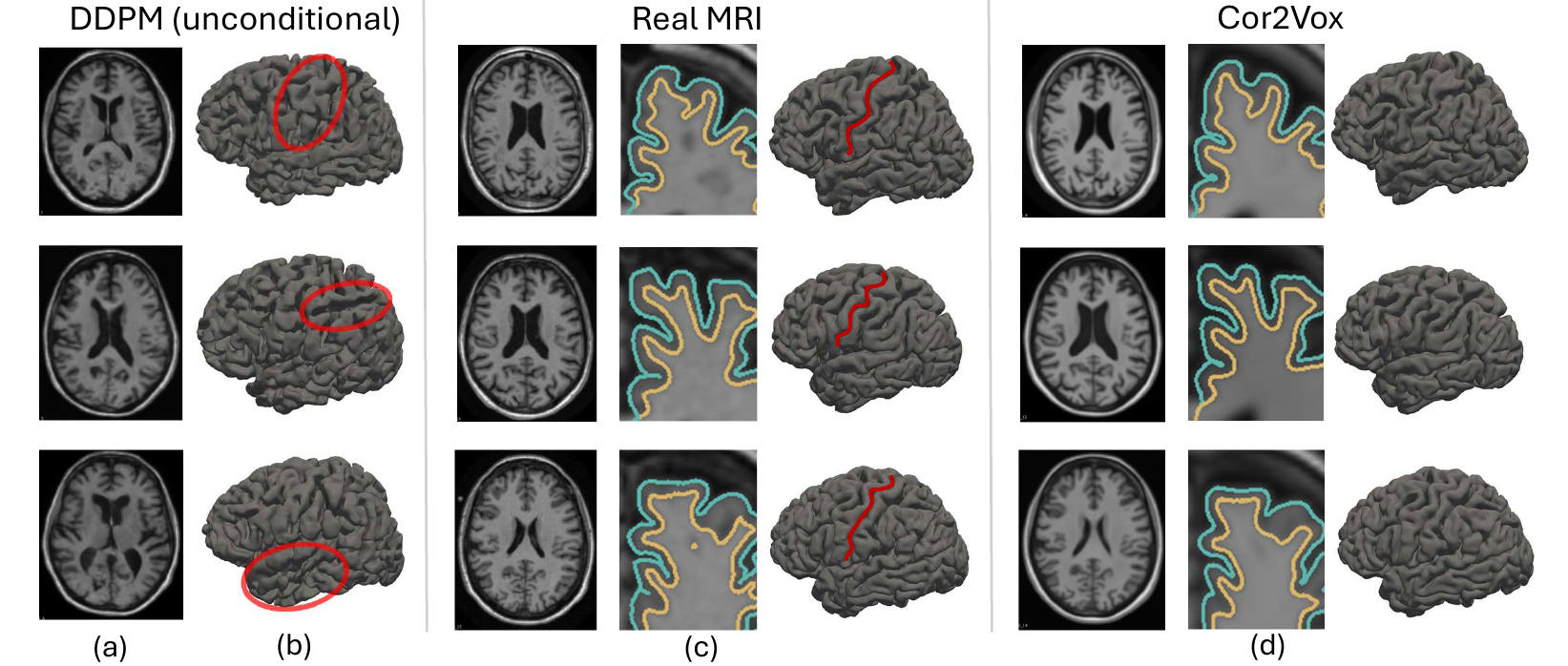}
    \caption{Conditioning on realistic cortical surfaces is crucial to avoid anatomical implausibilities in synthetic MRIs. (a) Generated MRIs from an unconditional 3D diffusion model. (b) Anatomical implausibilities; from top to bottom: missing central sulcus, unrealistic wide groove, scattered surface structure. (c) Real MRIs and corresponding cortical surfaces; we highlighted the characteristic central sulcus in red. (d) Our approach (Cor2Vox) incorporates real cortical surfaces to guide the generative process. 
    }
    \label{fig:motivation}
\end{figure}

Generating synthetic medical images has demonstrated numerous benefits, including improved model robustness~\cite{Gopinath2024synthetic}, enhanced fairness~\cite{burlina2021bias,Ktena2024generativefairness}, reduced manual annotation burdens~\cite{Menten2022simulation,Xanthis2021simulator}, and support in individualized simulations and treatment planning~\cite{zhu2008simulation}. 
Recent progress in diffusion models has enabled the generation of realistic-looking brain MRIs~\cite{Pinaya2022latentmri,Peng2023generatingbrainmri,Kim2024controllable} as in \Cref{fig:motivation}(a). 
However, examining beyond 2D views to analyze 3D cortical surface reconstructions reveals significant limitations, as most cases exhibit implausible folding patterns as shown in \Cref{fig:motivation}(b). We investigated 20 randomly generated MRIs and reconstructed their corresponding cortical surfaces of the left hemisphere. Severe irregularities were observed in 18 out of 20 reconstructions, including missing central gyri/sulci, unrealistic grooves, and scattered surface structures.
While initially surprising, this issue aligns with the well-established challenge diffusion models face in generating anatomically accurate hands \cite{narasimhaswamy2024handiffuser},
a problem that is further amplified given the geometric complexity of the cerebral cortex. 
This issue has likely received limited attention because traditional image quality metrics, such as Structural Similarity Index Measure (SSIM) and Peak Signal-to-Noise-Ratio
(PSNR), primarily evaluate visual similarity and fail to capture structural inconsistencies and critical neuroanatomical inaccuracies.

The problem of anatomical implausibility highlights a core challenge in image generation: achieving effective \emph{control} over the generative process.
While recent methods \cite{Zhang2023controlnet,mou2024t2i,zhao2024uni} have introduced localized conditions like 2D pose skeletons or edge maps to guide image synthesis, they are insufficient for the generation of brain MRIs, as controlling the intricate 3D geometry of the highly folded cortex demands more sophisticated representations.
Existing continuous shape representations, such as triangular meshes~\cite{Dale1999,MacDonald2000,Patenaude2011shapemodel} and signed distance fields (SDF)~\cite{Cruz2021deepcsr,gopinath2021segrecon,Han2004cruise}, excel at modeling the dense convolutions of the cortex and are instrumental in simulating sub-millimeter-level brain atrophy~\cite{Camara2006atrophy,Rusak2022benchmark}, aiding the benchmarking of segmentation and registration algorithms~\cite{CastellanoSmith2003simulation,Karacali2006simulation,Larson2022synthetic,Sharma2010atrophy}. 
Incorporating such representations into generative models therefore holds promise for synthesizing anatomically realistic 3D brain scans.

However, integrating geometric representations into standard denoising diffusion probabilistic models (DDPM)~\cite{ddpm,sohl-dickstein-diffusion2015} is suboptimal, as they assume a pure Gaussian noise as the prior, requiring anatomical constraints to be integrated externally. A recent method, Brownian Bridge diffusion models (BBDM)~\cite{li_bbdm_2023}, offers a more flexible approach by modeling the transport between two arbitrary distributions, allowing for a direct mapping between geometric conditions and brain MRIs.
Yet, current models~\cite{li_bbdm_2023,Lee2024ebdm} were developed for 2D images, and commonly operate in the latent space after a pre-trained autoencoder, making them unsuited for precisely generating intricate 3D structure of the cerebral cortex.

Thus, we introduce \emph{Cor2Vox}, the first diffusion model-based method leveraging a 3D shape-to-image Brownian bridge diffusion model for medical image synthesis incorporating continuous shape priors. 
We combine inner white matter and outer pial cortical surfaces deliberately to represent detailed brain structures with 3D SDFs. 
By leveraging both surfaces, Cor2Vox is tailored to model changes in cortical thickness, offering novel possibilities to accurately simulate cortical thinning. 
With the cortical control, we can compute surface distances to precisely quantify the accuracy of the generated tissue boundaries, surpassing the indirect evaluation with Cohen's d~\cite{Wu2024evaluating}. 
We show a significant improvement in the accuracy of the generated images over previous methods, while maintaining the model's ability to introduce variability in other image regions, such as the skull. 
Finally, we used Cor2Vox to simulate cortical atrophy in MRIs, supporting the benchmarking of cortical thickness estimation methods. 

\section{Related Work}
To date, generative adversarial networks (GAN)~\cite{goodfellow2014gan} and 
DDPMs~\cite{ddpm,sohl-dickstein-diffusion2015} are among the most prevalent approaches for image generation. 
Both have been extended to incorporate geometric and semantic constraints, such as segmentation masks and edge maps, to achieve more controlled  synthesis~\cite{pix2pix2017,Wang2018conditionalGAN,park2019SPADE,Rombach2022diffusion,Zhang2023controlnet}. 
However, these approaches are typically tailored to 2D images, making their adaptation to 3D medical data inherently challenging. 
Current approaches for 3D medical image generation commonly either
operate in a lower-dimensional latent space~\cite{Kim2024controllable,Peng2024metadataconditioned,Pinaya2022latentmri} or generate 3D volumes slice by slice~\cite{Han2023medgen3d,Peng2023generatingbrainmri,Li2024pasta}. Nonetheless, 
latent diffusion models have inherent limitations in achieving high precision~\cite{Konz2024controllableddpm,Rombach2022diffusion}, while slice-wise generative models introduce inter-slice inconsistencies which demand further post-processing~\cite{Han2023medgen3d,Li2024pasta}.
Recent advancements, such as Med-DDPM~\cite{Dorjsembe2024medddpm}, directly generate 3D medical scans conditioned on voxel segmentation masks, providing control over the location and shape of structures like tumors~\cite{Dorjsembe2024medddpm,Kim2024controllable}. 
Yet, their level of detail is limited by the voxel size, 
which poses challenges when generating fine details for complex geometries, such as the cortex.
While this discussion focuses on geometric condition-based image generation, it is worth noting that alternative approaches for synthesizing medical images exist as well, including methods based on Jacobian deformations~\cite{Karacali2006simulation,Sharma2010atrophy} and biophysical models~\cite{Camara2006atrophy,CastellanoSmith2003simulation,Khanal2016biophysical}.

\section{Methods}

\subsection{Cortical Shape Representations}

Triangular surface meshes are a common \emph{explicit} representation of cortical tissue boundaries~\cite{fischlFreeSurfer2012,Bongratz2024v2cflow}, directly usable for visualization and rendering. 
Alternatively, surfaces can be represented \emph{implicitly} through signed distance fields (SDF). 
Given a surface $\mathcal{S}\subset\mathbb{R}^3$, an SDF is defined as a function: 
\begin{equation} 
g: \Omega \subseteq \mathbb{R}^3 \to \mathbb{R}, 
\end{equation} 
which maps each point $x\in\mathbb{R}^3$ to its orthogonal distance to the surface $\mathcal{S}$. 
The zero-level set of $g$ corresponds to the surface $\mathcal{S}$, and the sign of $g$ indicates whether $x$ lies inside ($g(x)<0$) or outside ($g(x)>0$) the surface.
With distance transforms~\cite{Danielsson1980distance}, surface meshes can be converted into SDFs by computing the orthogonal distance from an input point to its closest triangle. Hence, SDFs can be represented as dense voxel grids, making them compatible with 3D image data. Additionally, meshes can be converted into binary edge maps by identifying the occupancy of voxels with mesh faces. 


\begin{figure}[t]
    \centering
    \includegraphics[width=0.9\linewidth]{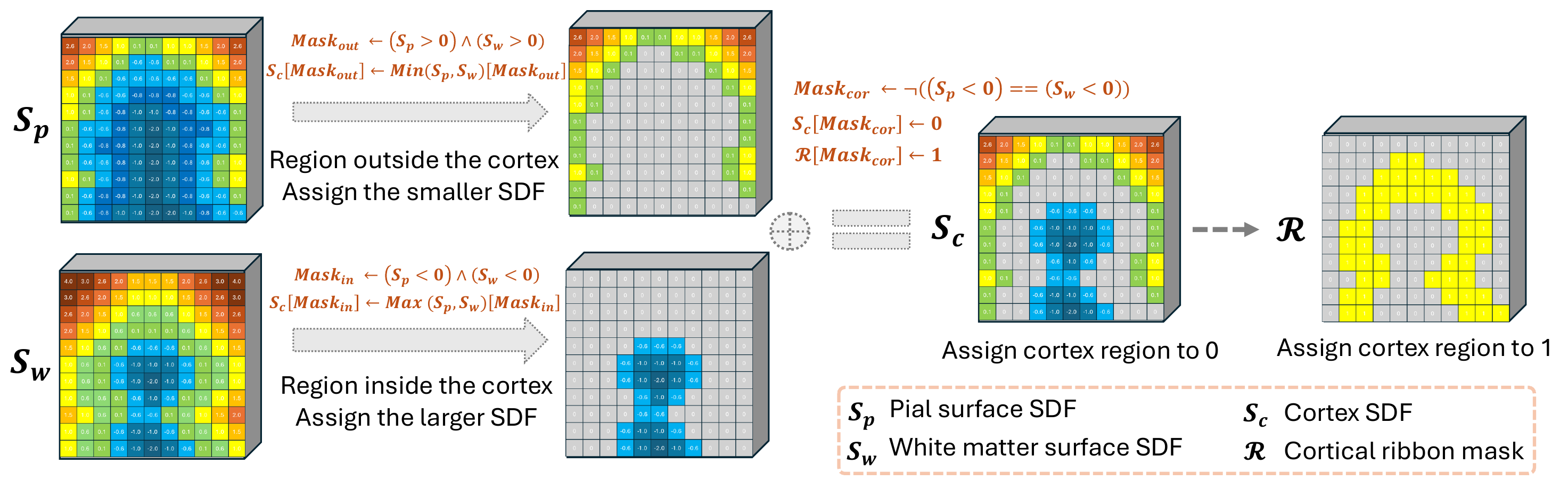}
    \caption{Generation of cortex SDF ($\mathcal{S}_c$) and cortical ribbon mask ($\mathcal{R})$.}
    \label{fig:cortexsdf}
\end{figure}

As shown in \Cref{fig:motivation}(c), the cerebral cortex has outer and inner surface boundaries, namely the pial and white matter surface, which are commonly represented as separate meshes $\mathcal{M}_p$ and $\mathcal{M}_w$, respectively. 
We aim to combine both surfaces into a single cortex-describing SDF, $\mathcal{S}_c$, which will be the source input to our framework.
To achieve this, we first convert the cortical meshes into dense SDFs using a distance transform, yielding $\mathcal{S}_p$ and $\mathcal{S}_w$.
Subsequently, we construct the cortex SDF $\mathcal{S}_c$ by combining $\mathcal{S}_p$ and $\mathcal{S}_w$ 
following the procedure in~\Cref{fig:cortexsdf}. We begin by initializing $\mathcal{S}_c$ as a tensor of the same size as $\mathcal{S}_p$ and $\mathcal{S}_w$, with all entries set to zero. The regions of interest are then identified and processed as follows. 
i) The region outside the cortex is identified where $\mathcal{S}_p$ and $\mathcal{S}_w$ have both positive values; within this region, we assign the lower value, i.e., the distance to the pial surface, to $\mathcal{S}_c$.
ii) Next, the region inside the cortical ribbon is determined as $\mathcal{S}_p$ and $\mathcal{S}_w$ having both negative values; within this region, we assign the higher value, i.e., the distance to the white matter surface, to $\mathcal{S}_c$. 
iii) Lastly, the cortical ribbon is given by the region where $\mathcal{S}_p$ and $\mathcal{S}_w$ have opposite signs; we set $\mathcal{S}_c$ in this region to zero to ensure a clear cortical boundary. We also extract this cortical ribbon as a separate mask $\mathcal{R}$.

\begin{figure}[t]
    \centering
    \includegraphics[width=0.9\linewidth]{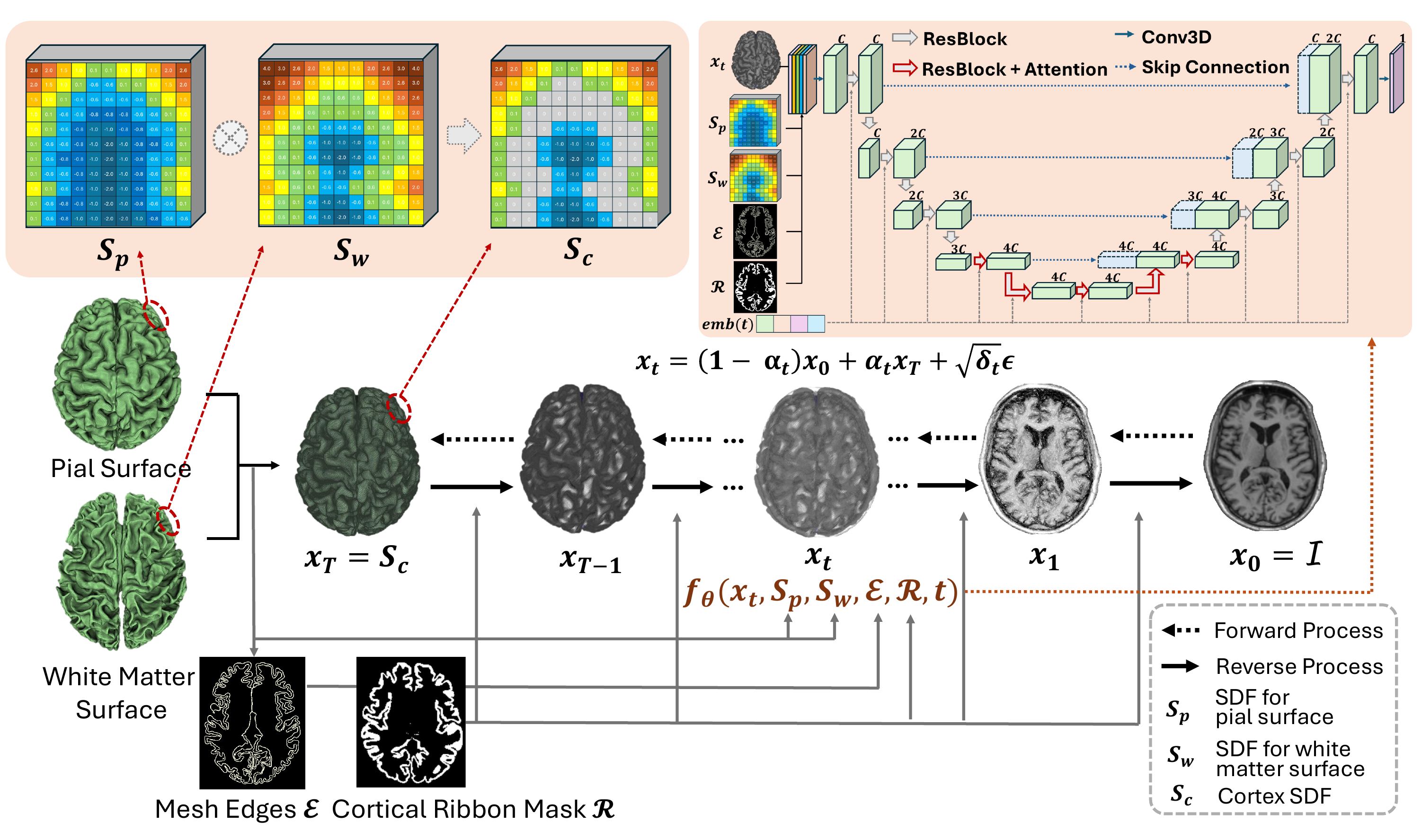}
    \caption{Cor2Vox leverages a shape-to-image Brownian bridge diffusion process to learn a stochastic mapping $f_\theta$ between the shape prior $\mathcal{S}_c$ and the MRI domain $\mathcal{I}$. During the reverse diffusion process, additional shape conditions are incorporated to improve structural alignment in the generated MRIs. $f_\theta$ is modeled using a 3D UNet.}
    \label{fig:architecture}
\end{figure}

\subsection{Shape-to-Image Brownian Bridge Diffusion}
Our objective is to generate an image $\mathcal{I} = f_\theta(\mathcal{S}_c)$ that precisely matches the geometry given by a shape prior $\mathcal{S}_c$.
Inspired by \cite{li_bbdm_2023}, we use a Brownian bridge process and adapt it for directly transforming the shape to the 3D image domain. We show an overview of our method, \emph{Cor2Vox}, in \Cref{fig:architecture}.
Cor2Vox uses paired data $(\mathcal{S}_c^i, ~\mathcal{I}^i)_{i=1}^N$ for training, comprising $N$ pairs of 3D MRIs $\mathcal{I}\in{\mathbb{R}^{H\times W\times D}}$ and corresponding cortical shapes. We assume that shapes are given as dense SDFs $\mathcal{S}_c\in{\mathbb{R}^{H\times W\times D}}$ with the same dimensionality as the MRIs.

\noindent
\textbf{Forward Process.} Unlike DDPM, which perturbs the input until pure Gaussian noise $\varepsilon \sim \mathcal{N}(0, \mathbf{I}$) at timestep $T$, the Brownian bridge 
diffusion process maps between 
data points from the joint distribution of structured  
source and target domains, i.e., $(\boldsymbol{x}_T, \boldsymbol{x}_0) \sim q_{data}(\mathcal{S}_c, ~\mathcal{I})$. Starting from an initial state $\boldsymbol{x}_0$ (i.e., the MRI volume $\mathcal{I}$), the forward process of the Brownian bridge connects it to the destination state $\boldsymbol{x}_T$ (i.e., the shape prior $\mathcal{S}_c$) by:
\begin{equation}
    q(\boldsymbol{x}_t \mid \boldsymbol{x}_0, \boldsymbol{x}_T) = \mathcal{N}\big(\boldsymbol{x}_t; (1 - \alpha_t)\boldsymbol{x}_0 + \alpha_t \boldsymbol{x}_T, \delta_t \mathbf{I}\big), \quad \text{where} \quad \mathcal{S}_c = \boldsymbol{x}_T,
\end{equation}

\noindent
$\alpha_t = t/T$, $T$ the total number of steps in the diffusion process, and $\delta_t = 2(\alpha_t - {\alpha_t}^2)$ the variance term. The transition probability between two consecutive steps can be derived as~\cite{li_bbdm_2023}:
\begin{align}
\label{eq:bbdm_forward}
    q_{\text{BB}}(\boldsymbol{x}_t \mid \boldsymbol{x}_{t-1}, \boldsymbol{x}_T) &= \mathcal{N}\bigg(\boldsymbol{x}_t; \frac{1 - \alpha_t}{1 - \alpha_{t-1}} \boldsymbol{x}_{t-1} + \left(\alpha_t - \frac{1 - \alpha_t}{1 - \alpha_{t-1}} \alpha_{t-1}\right) \boldsymbol{x}_T, \delta_{t \mid t-1} \mathbf{I}\bigg), \nonumber \\
    \delta_{t \mid t-1} &= \delta_t - \delta_{t-1} \frac{(1 - \alpha_t)^2}{(1 - \alpha_{t-1})^2}.
\end{align}

\begin{algorithm}[t]
\caption{Training Process}
\label{alg:training}
\begin{algorithmic}[1]
\Repeat
    \State Paired data: $\boldsymbol{x}_0 \sim q(\mathcal{I}), \boldsymbol{x}_T \sim q(\mathcal{S}_c), \mathcal{C} = (\mathcal{S}_p, \mathcal{S}_w, \mathcal{E}, \mathcal{R}) \sim q(\mathcal{S}_p, \mathcal{S}_w, \mathcal{E}, \mathcal{R})$
    \State Timestep $t \sim \text{Uniform}(1, \dots, T)$
    \State Gaussian noise $\varepsilon \sim \mathcal{N}(0, \mathbf{I})$
    \State Forward diffusion $\boldsymbol{x}_t = (1 - \alpha_t)\boldsymbol{x}_0 + \alpha_t \boldsymbol{x}_T + \sqrt{\delta_t} \varepsilon$
    \State Take gradient descent step on
    $\nabla_\theta 
    \lVert \alpha_t (\boldsymbol{x}_T - \boldsymbol{x}_0) 
    + \sqrt{\delta_t} \boldsymbol{\varepsilon} - f_\theta(\boldsymbol{x}_t, \mathcal{C}, t) \rVert_1$
\Until{converged}
\end{algorithmic}
\end{algorithm}

\begin{algorithm}[t]
\caption{Sampling Process}
\begin{algorithmic}[1]
\State Sample conditional input $x_T = \mathcal{S}_c \sim q(\mathcal{S}_c)$, $\mathcal{C} = (\mathcal{S}_p, \mathcal{S}_w, \mathcal{E}, \mathcal{R}) \sim q(\mathcal{S}_p, \mathcal{S}_w, \mathcal{E}, \mathcal{R})$
\For{$t = T, \dots, 1$}
    \State \textbf{if} $t > 1$ \textbf{then} $\varepsilon \sim \mathcal{N}(0, \mathbf{I})$, \textbf{else} $\varepsilon = 0$
    \State $x_{t-1} = c_{xt} \boldsymbol{x}_t + c_{st} \mathcal{S}_c - c_{ft} f_\theta \big(\boldsymbol{x}_t, \mathcal{C}, t \big) + \sqrt{\tilde{\delta}_t} \varepsilon$
\EndFor
\State \Return $x_0$
\end{algorithmic}
\label{alg:sampling}
\end{algorithm}

Thus, at the beginning of the diffusion process, i.e., $t = 0$, we have $\alpha_0 = 0$, starting from the mean value of $\boldsymbol{x}_0 = \mathcal{I}$ with probability 1 and variance $\delta_0 = 0$.
As the diffusion progresses, the variance $\delta_t$ first increases to its maximum value $\delta_{max} = \delta_{T/2} = 1/2$ at the midpoint of the process, and then decreases until it reaches $\delta_{T} = 0$ at $\alpha_T = 1$. At this point, the diffusion concludes in the destination with a mean value equal to $\mathcal{S}_c$. Through this mechanism, the forward diffusion process establishes a stochastic mapping with fixed endpoints from the image domain $\mathcal{I}$ to the shape domain $\mathcal{S}_c$.  \\

\noindent
\textbf{Reverse Process.} The reverse process of BBDM is designed to predict $\boldsymbol{x}_{t-1}$ given $\boldsymbol{x}_t$, starting directly with the conditional input $\boldsymbol{x}_T = \mathcal{S}_c$, different from standard diffusion models that start from the pure Gaussian noise. To improve the structural alignment, we incorporate supplementary shape representations that can be obtained from the cortical meshes. This includes the SDF of the pial surface $\mathcal{S}_p$, the SDF of the white matter surface $\mathcal{S}_w$, a binary edge map $\mathcal{E}$ containing both surfaces, and the cortical ribbon mask $\mathcal{R}$, cf.~\Cref{fig:cortexsdf}. We collectively represent these conditions as $\mathcal{C} = (\mathcal{S}_p, \mathcal{S}_w, \mathcal{E}, \mathcal{R})$, and integrate them at each timestep $t$ to aid the prediction:
\begin{equation}
\label{eq:bbdm_reverse}
    p_\theta(\boldsymbol{x}_{t-1} \mid \boldsymbol{x}_t, \mathcal{C}, \mathcal{S}_c) = \mathcal{N}\big(\boldsymbol{x}_{t-1}; \boldsymbol{\mu}_\theta(\boldsymbol{x}_t, \mathcal{C}, t), \tilde{\delta}_t \mathbf{I}\big).
\end{equation}
\noindent Here, $\boldsymbol{\mu}_\theta(\boldsymbol{x}_t, \mathcal{C}, t)$ represents the predicted mean value while $\tilde{\delta}_t$ denotes the variance of noise at each timestep. Following the reparameterization strategy used in DDPM~\cite{ddpm}, we can train a neural network $f_\theta(\cdot)$ to predict solely the noise instead of the mean $\boldsymbol{\mu}_\theta$. Namely, we reformulate $\boldsymbol{\mu}_\theta$ as a linear combination of $\boldsymbol{x}_t$,  $\mathcal{S}_c$, and the estimated part $f_\theta$:
\begin{align}
\label{eq:bbdm_ep}
\boldsymbol{\mu}_\theta(\boldsymbol{x}_t, \mathcal{C}, \mathcal{S}_c, t) &= c_{xt} \boldsymbol{x}_t + c_{st} \mathcal{S}_c + c_{ft} f_\theta(\boldsymbol{x}_t, \mathcal{C}, t), \, \text{where} \\
    c_{xt} &= \frac{\delta_{t-1}}{\delta_t} \frac{1 - \alpha_t}{1 - \alpha_{t-1}} + \frac{\delta_{t|t-1}}{\delta_t} (1 - \alpha_{t-1}), \nonumber  \\
    c_{st} &= \alpha_{t-1} - \alpha_t \frac{1 - \alpha_t}{1 - \alpha_{t-1}} \frac{\delta_{t-1}}{\delta_t}, \nonumber \, \text{and}  \\
    c_{ft} &= (1 - \alpha_{t-1}) \frac{\delta_{t|t-1}}{\delta_t}. \nonumber
\end{align}
The parameters $c_{xt}$, $c_{st}$, and $c_{ft}$ are non-trainable. Instead, they are derived from $\alpha_t$, $\alpha_{t-1}$, $\theta_t$, and $\theta_{t-1}$.
The variance term $\tilde{\delta}_t$ does not need to be learned either; it can be derived in the analytical form as $\tilde{\delta}_t = \frac{\delta_{t|t-1} \cdot \delta_{t-1}}{\delta_t}$. \\

\noindent
\textbf{Training.} We outline the training process in \Cref{alg:training}. The training is designed to minimize the disparity between the joint distribution predicted by the model $f_\theta$ and the training data. This is achieved by optimizing the Evidence Lower Bound (ELBO) defined below:
\begin{align}
\text{ELBO} &= - \mathbb{E}_q \big( \text{D}_{\text{KL}}(q_{\text{BB}}(\boldsymbol{x}_T \vert ~\boldsymbol{x}_0, \mathcal{S}_c) \parallel p(\boldsymbol{x}_T \vert ~\mathcal{C}, \mathcal{S}_c))  \nonumber \\
&\quad + \sum_{t=2}^T \text{D}_{\text{KL}}(q_{\text{BB}}(\boldsymbol{x}_{t-1} \vert ~\boldsymbol{x}_t, \boldsymbol{x}_0, \mathcal{S}_c) \parallel p_\theta(\boldsymbol{x}_{t-1} \vert ~\boldsymbol{x}_t, \mathcal{C}, \mathcal{S}_c)) \nonumber \\
&\quad - \log p_\theta(\boldsymbol{x}_0 \vert ~\boldsymbol{x}_1, \mathcal{C}, \mathcal{S}_c) \big).
\end{align}
By combining the ELBO with~\Cref{eq:bbdm_forward,eq:bbdm_reverse,eq:bbdm_ep}, the training objective is given by:
\begin{equation}
    \mathcal{L}_{\text{Cor2Vox}} =
    \mathbb{E}_{x_0, \mathcal{S}_c, \varepsilon \sim \mathcal{N}(0, \mathbf{I})} \Big[
    \Big\lVert \alpha_t (\mathcal{S}_c - \boldsymbol{x}_0) 
    + \sqrt{\delta_t} \boldsymbol{\varepsilon} - f_\theta(\boldsymbol{x}_t, \mathcal{C}, t) \Big\rVert_1
    \Big].
\end{equation}
\\

\noindent
\textbf{Sampling.} We outline the sampling process in \Cref{alg:sampling}. The sampling process can be derived as:
\begin{equation}
    \boldsymbol{x}_{t-1} = c_{xt} \boldsymbol{x}_t + c_{st} \mathcal{S}_c - c_{ft} f_\theta \big(\boldsymbol{x}_t, \mathcal{C}, t \big) + \sqrt{\tilde{\delta}_t} \varepsilon,
\end{equation}
\noindent where $\varepsilon \sim \mathcal{N}(0, \mathbf{I})$ when $t > 1$, otherwise $\varepsilon = 0$. We accelerate the sampling process using the DDIM~\cite{ddim} strategy, which adopts a non-Markovian process with the same marginal distributions as Markovian inference.

\subsection{Surface-Based Evaluation of Geometric Accuracy}
Since we use continuous brain surfaces, $\mathcal{M}^{\text{Ref}}$, as the input to our model, we can evaluate the geometric accuracy of the generated images via surface-based distance metrics. This is not possible for voxel-based methods, where no such reference is available for comparison. 
Specifically, we propose to use V2C-Flow~\cite{Bongratz2024v2cflow} to obtain cortical surfaces, $\mathcal{M}^{\text{Pred}}$ from the generated images. Based on these reconstructions, we compute the average symmetric surface distance (ASSD):
\begin{equation}
  \mathrm{ASSD}(\mathcal{M}^{\text{Pred}}, \mathcal{M}^{\text{Ref}}) = 
  \frac{\sum_{p \in \mathcal{P}^{\text{Pred}}} d(p,\mathcal{M}^{\text{Ref}}) + \sum_{p \in \mathcal{P}^{\text{Ref}}} d(p,\mathcal{M}^{\text{Pred}})}
  {|\mathcal{P}^{\text{Pred}}| + |\mathcal{P}^{\text{Ref}}|},
\end{equation}
using $|\mathcal{P}^{\text{Pred}}|=|\mathcal{P}^{\text{Ref}}|=$100,000 randomly sampled surface points. The distance $d(p, \mathcal{M})$ measures the orthogonal distance from a point $p\in \mathbb{R}^3$ to its closest triangle in the mesh $\mathcal{M}$.



\section{Results}

\subsection{Experimental Setting}

\noindent
\textbf{Models and hyperparameters.} We employ a 3D UNet to model $f_\theta$, which is a 3D adaptation of the ADM architecture~\cite{DMbeatsGAN}. 
We use channels of $[C, 2C, 3C, 4C]$ for each residual stage, where $C = 64$. Global attention is applied at downsampling factor 8, with 4 heads and 64 channels. We use adaptive group normalization to inject the timestep embedding into each residual block.
The model is trained using the Adam optimizer with an initial learning rate of $1 \times 10^{-4}$, reduced by a factor of $0.5$ on the plateau, a batch size of $2$, and an exponential moving average (EMA) rate of $0.995$, for 400 epochs with one NVIDIA H100 GPU. We set 1,000 timesteps for training, while 10 for inference with DDIM~\cite{ddim} sampling strategy for a balance between image quality and computational efficiency.
\\

\begin{figure}[t]
    \centering
    \includegraphics[width=\linewidth]{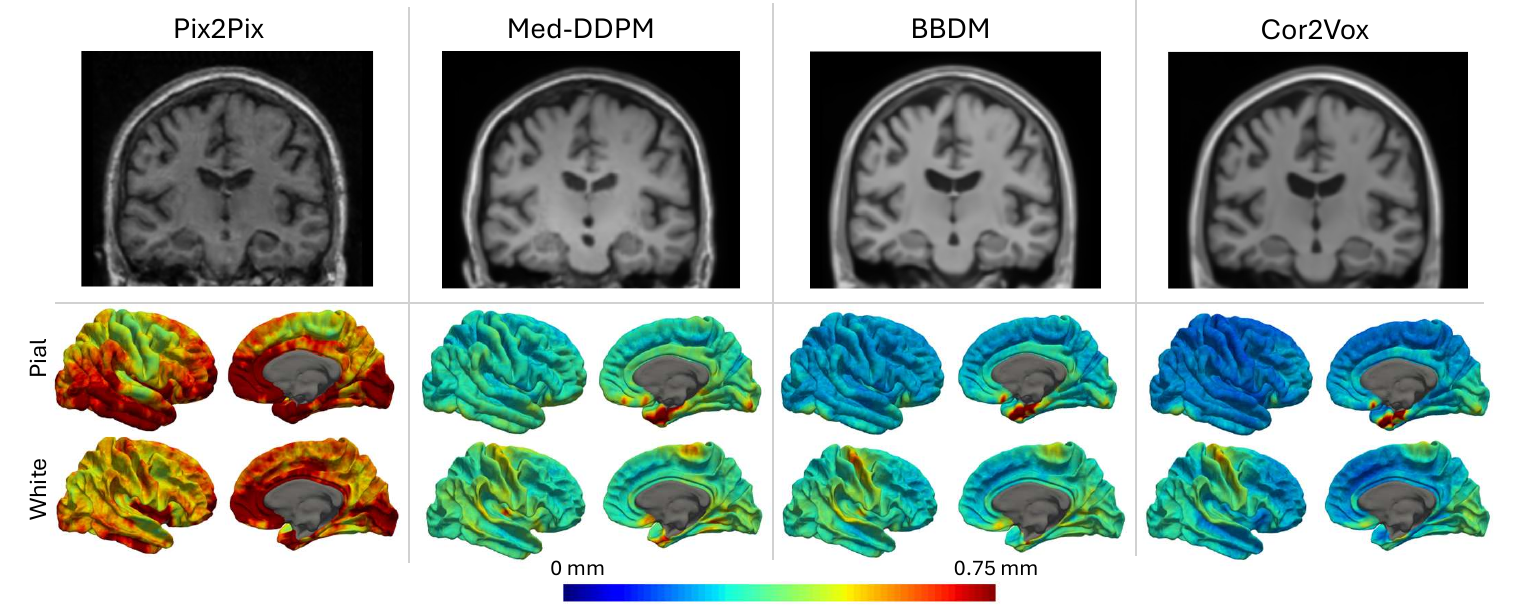}
    \caption{Generated brain MRIs (top) and vertex-wise errors (bottom, in mm) comparing reconstructed cortical shapes to their original inputs. We show the mean error values on the cortical surface across the test set for the right hemisphere, separately for pial and white matter surfaces.}
    \label{fig:vv-error}
\end{figure}

\noindent
\textbf{Data and preprocessing.}
We used data from the Alzheimer's Disease Neuroimaging Initiative~(ADNI, \url{https://adni.loni.usc.edu}), containing T1-weighted brain MRIs from cognitively normal subjects, subjects diagnosed with mild cognitive impairment~(MCI) and Alzheimer's disease~(AD). We registered all scans to MNI152 space via affine registration, implemented in NiftyReg (v1.5.69). We also processed all data with FreeSurfer (v7.2)~\cite{fischlFreeSurfer2012}. To avoid subject bias, we used only baseline scans, with 1,155/169/323 samples for train/validation/test sets. 
We used the pre-trained V2C-Flow~\cite{Bongratz2024v2cflow} model to extract surfaces as input to Cor2Vox for both training and sampling. 
From the surface meshes, we created white matter and pial SDFs, merging both hemispheres into one SDF, with a KDTree~\cite{Maneewongvatana2002kdtree} implemented in Scipy~(v1.10.0). Internally, our models use a resolution of $128^3$ voxels but we rescaled all outputs to MNI space (1mm isotropic resolution, trilinear interpolation) for evaluation. To assess image quality, we used SynthStrip~\cite{Hoopes2022synthstrip} to skull-strip both generated and original images (\texttt{orig.mgz} files from FreeSurfer), to avoid the scanner noise in the original data to confound the evaluation. Nonetheless, all models were trained to generate the entire MRI, i.e., including the skull and background, as shown in our visualizations.
\\






\noindent
\textbf{Baselines implementation.}
We compare Cor2Vox to state-of-the-art conditional image generation models, namely Pix2Pix~\cite{pix2pix2017}, Med-DDPM~\cite{Dorjsembe2024medddpm}, and BBDM~\cite{li_bbdm_2023}. 
We adapted Pix2Pix and BBDM for 3D data as both were originally developed for 2D images. 
For BBDM, we applied the Brownian diffusion process in the image space instead of the latent space, in analogy to Med-DDPM and Cor2Vox.  
We adapted Med-DDPM by switching the intensity scaling from slice-wise to volumetric and the denoising target from noise to denoised image, which considerably enhanced its performance. 
We provided the mesh edges ($\mathcal{E}$) and cortical ribbon masks ($\mathcal{R}$) as input conditions to these models, as they closely resemble the masks and edge maps used in the original works.
In addition, we developed a variant of Cor2Vox by replacing the Brownian bridge process with a standard denoising diffusion process with the same input as Cor2Vox; we call this variant Cor2Vox{\scriptsize/DDPM}.

\begin{table}[t]
    \setlength{\tabcolsep}{1.8pt}
    \renewcommand\bfdefault{b}
    \centering
    \caption{Quantitative comparison of implemented methods for 3D brain MRI generation. We report mean \textpm~SD across all samples in our test set. Geometric accuracy of the cortex is in mm.}
    \begin{threeparttable}
    \begin{tabular}{lcccccc}
    \toprule
    &&& \multicolumn{4}{c}{Geometric Accuracy --- ASSD$\downarrow$} \\
    \cmidrule(lr){4-7}
    & \multicolumn{2}{c}{Image Quality} & \multicolumn{2}{c}{Left Hemisphere} & \multicolumn{2}{c}{Right Hemisphere} \\
    \cmidrule(lr){2-3}
    \cmidrule(lr){4-5}
    \cmidrule(lr){6-7}
    Model 
    & SSIM$\uparrow$ & PSNR$\uparrow$ & White & Pial & White & Pial \\

    \midrule
    
    Pix2Pix$_\mathcal{E}$\cite{pix2pix2017}\tnote{*}
    & 0.891{\tiny\textpm0.011}
    & \textbf{24.76{\tiny\textpm1.96}}
    & 0.784{\tiny\textpm0.190}
    & 0.853{\tiny\textpm0.227}
    & 0.760{\tiny\textpm0.174}
    & 0.804{\tiny\textpm0.195}
    \\

    Med-DDPM$_\mathcal{E}$\cite{Dorjsembe2024medddpm}
    & 0.894{\tiny\textpm0.015}
    & 22.38{\tiny\textpm2.63}
    & 0.506{\tiny\textpm0.042}
    & 0.437{\tiny\textpm0.027}
    & 0.516{\tiny\textpm0.043}
    & 0.436{\tiny\textpm0.026}
    \\

    BBDM$_\mathcal{E}$\cite{li_bbdm_2023}\tnote{*}
    & 0.898{\tiny\textpm0.020}
    & 19.95{\tiny\textpm2.61}
    & 0.349{\tiny\textpm0.032}
    & 0.307{\tiny\textpm0.021}
    & 0.368{\tiny\textpm0.035}
    & 0.307{\tiny\textpm0.021}
    \\
    
    \midrule

    Pix2Pix$_\mathcal{R}$\cite{pix2pix2017}\tnote{*}
    & 0.885{\tiny\textpm0.013}
    & 23.69{\tiny\textpm2.06}
    & 0.697{\tiny\textpm0.137}
    & 0.691{\tiny\textpm0.155}
    & 0.698{\tiny\textpm0.183}
    & 0.680{\tiny\textpm0.221}
    \\


    Med-DDPM$_\mathcal{R}$\cite{Dorjsembe2024medddpm}
    & 0.899{\tiny\textpm0.016}
    & 22.30{\tiny\textpm2.80}
    & 0.359{\tiny\textpm0.043}
    & 0.347{\tiny\textpm0.034}
    & 0.373{\tiny\textpm0.106}
    & 0.352{\tiny\textpm0.101}
    \\


    BBDM$_\mathcal{R}$\cite{li_bbdm_2023}\tnote{*}
    & 0.898{\tiny\textpm0.021}
    & 19.67{\tiny\textpm2.54}
    & 0.328{\tiny\textpm0.037}
    & 0.283{\tiny\textpm0.034}
    & 0.338{\tiny\textpm0.128}
    & 0.287{\tiny\textpm0.121}
    \\






    
    \midrule

    Cor2Vox{\scriptsize/DDPM}
    & 0.902{\tiny\textpm0.015}
    & 23.13{\tiny\textpm2.71}
    & 0.335{\tiny\textpm0.038}
    & 0.300{\tiny\textpm0.033}
    & 0.348{\tiny\textpm0.048}
    & 0.305{\tiny\textpm0.037}
    \\
    
    Cor2Vox
    & \textbf{0.906{\tiny\textpm0.018}}
    & 21.10{\tiny\textpm2.74}
    & \textbf{0.283{\tiny\textpm0.029}}
    & \textbf{0.251{\tiny\textpm0.019}}
    & \textbf{0.289{\tiny\textpm0.031}}
    & \textbf{0.251{\tiny\textpm0.017}}
    \\

    \bottomrule
    
    \end{tabular}

    \begin{tablenotes} \scriptsize
            \item[*] Method adapted for 3D image generation.
        \end{tablenotes}
    \end{threeparttable}
    \label{tab:qual-acc}
\end{table}

\subsection{Image Quality and Accuracy}
We visualize the generated brain MRIs from Cor2Vox and implemented baseline methods in \Cref{fig:vv-error}. Additionally, we show local surface-based errors on the FsAverage template that serves as input to V2C-Flow. We report quantitative scores for all methods in \Cref{tab:qual-acc}. Note that the reconstruction errors shown in \Cref{fig:vv-error} are not equivalent to the ASSD; the ASSD is bi-directional and independent of vertices, whereas the surface plots show the average distance of vertices in the reconstructed surfaces to the original cortical boundaries. 

From the qualitative inspection of the generated images, we observe that all methods are capable of generating the same cortical anatomy based on the provided shape condition.
However, the quantitative evaluation in~\Cref{tab:qual-acc} reveals a clear improvement of Cor2Vox over other methods in geometric accuracy, confirmed by surface-based plots in \Cref{fig:vv-error}. 
White matter surfaces are, on average, by approximately 0.04\,mm ($\sim$10\%) more accurate than those from the 3D BBDM. For the pial surfaces, the gain is similar with a reduced error of around 0.03\,mm. At the same time, Cor2Vox does not sacrifice image quality and achieves the highest SSIM score, albeit only by a slight margin. The 3D Pix2Pix excels in terms of PSNR, however, with an ASSD of more than 0.5\,mm on all surfaces, Pix2Pix is not competitive regarding reconstruction accuracy. Med-DDPM and the Cox2Vox{\scriptsize /DDPM} variant also achieve low errors on both white and pial surfaces but cannot keep up with the best Brownian bridge-based methods. Generally, the surface errors are slightly larger on the white matter than on the pial surfaces, especially in the precentral gyrus. This could be explained by the challenge of precisely synthesizing the subtle contrast between white and gray matter, compared to the more pronounced intensity difference between gray matter and cerebrospinal fluid. Yet, Cor2Vox is the only method achieving an ASSD below 0.3\,mm for the white matter surface.
Finally, paired Wilcoxon signed-rank tests between Cor2Vox and baseline methods indicated highly significant improvements with $p<10^{-7}$ for the geometric accuracy on both surfaces and hemispheres.

\subsection{Ablation Study}

We evaluate the impact of various shape conditions, whether as the source domain for the generative process or as additional conditioning inputs in Cor2Vox, on the geometric accuracy. As reported in~\Cref{tab:ablation}, we initially tested individual shape conditions --- pial surface SDF $\mathcal{S}_p$, white matter surface SDF $\mathcal{S}_w$, edge map $\mathcal{E}$, cortical ribbon mask $\mathcal{R}$, cortex SDF $\mathcal{S}_c$ --- as single-source domains in the Brownian bridge diffusion process. The pial surface achieves the best accuracy when conditioned solely on $\mathcal{S}_p$, and similarly for the white matter surface with $\mathcal{S}_w$. The cortical ribbon mask ($\mathcal{R}$) helps with the accuracy of the white matter while falling short on the pial surface. Next, we explored using $\mathcal{R}$ as the source domain and its combination with other conditions. This approach performs better than either using no additional conditions or employing two parallel bridge processes for both $\mathcal{S}_p$ and $\mathcal{S}_w$. Finally, we adopt the cortex SDF $\mathcal{S}_c$ as the source domain and experiment with incorporating different sets of shape conditions as additional inputs during the reverse diffusion process. Results show that including more conditions generally improves performance, with the combination of all four shape conditions yielding the best accuracy on the white matter surface and comparable performance on the pial surface across hemispheres. This validates the effectiveness of Cor2Vox with $\mathcal{S}_c$ as the source domain and the importance of including extra shape information via conditioning.

\begin{table}[t]
    \setlength{\tabcolsep}{7pt}
    \renewcommand\bfdefault{b}
    \centering
    \caption{Ablation study of different source domains for the Brownian bridge process and extra input conditions provided to Cor2Vox. We report the mean\textpm SD on the validation set. The geometric accuracy of the cortex is in mm. 
    ($\mathcal{S}_p$: Pial SDF, $\mathcal{S}_w$: White matter SDF, $\mathcal{S}_c$: Cortex SDF, $\mathcal{E}$: Edge map, $\mathcal{R}$: Cortical ribbon mask)}
    \begin{threeparttable}
        
    \begin{tabular}{llcccccc}
    \toprule
    && \multicolumn{4}{c}{Geometric Accuracy --- ASSD$\downarrow$} \\
    \cmidrule(lr){3-6}
    && \multicolumn{2}{c}{Left Hemisphere} & \multicolumn{2}{c}{Right Hemisphere} \\
    \cmidrule(lr){3-4}
    \cmidrule(lr){5-6}
    Source & Condition 
    & White & Pial & White & Pial \\
    \midrule
  
    $\mathcal{S}_p$ & ---
    & 0.375{\tiny\textpm0.066}
    & \textbf{0.233{\tiny\textpm0.021}}
    & 0.386{\tiny\textpm0.068}
    & \textbf{0.236{\tiny\textpm0.022}}
    \\

    $\mathcal{S}_w$ & ---
    & \underline{0.289{\tiny\textpm0.066}}
    & 0.375{\tiny\textpm0.031}
    & 0.303{\tiny\textpm0.067}
    & 0.376{\tiny\textpm0.034}
    \\


    
    $\mathcal{E}$ & ---
    & 0.351{\tiny\textpm0.037}
    & 0.307{\tiny\textpm0.023}
    & 0.369{\tiny\textpm0.042}
    & 0.307{\tiny\textpm0.024}
    \\

    $\mathcal{R}$ & ---
    & 0.336{\tiny\textpm0.034}
    & 0.279{\tiny\textpm0.021}
    & 0.335{\tiny\textpm0.037}
    & 0.280{\tiny\textpm0.022}
    \\

    


    $\mathcal{S}_c$ & ---
    & 0.381{\tiny\textpm0.066}
    & 0.245{\tiny\textpm0.026}
    & 0.390{\tiny\textpm0.070}
    & 0.247{\tiny\textpm0.026}
    \\
    \midrule

    $\mathcal{S}_p, \mathcal{S}_w$\tnote{*} & ---
    & 0.394{\tiny\textpm0.074}
    & 0.535{\tiny\textpm0.064}
    & 0.418{\tiny\textpm0.079}
    & 0.553{\tiny\textpm0.071}
    \\

    $\mathcal{R}$ & $\mathcal{S}_p, \mathcal{S}_w$
    & 0.293{\tiny\textpm0.027}
    & 0.262{\tiny\textpm0.018}
    & 0.305{\tiny\textpm0.031}
    & 0.261{\tiny\textpm0.016}
    \\

    $\mathcal{R}$  & $\mathcal{S}_p, \mathcal{S}_w, \mathcal{E}$
    & 0.301{\tiny\textpm0.038}
    & 0.263{\tiny\textpm0.021}
    & 0.305{\tiny\textpm0.040}
    & 0.264{\tiny\textpm0.022}
    \\
    \midrule

    $\mathcal{S}_c$  & $\mathcal{E}$
    & 0.326{\tiny\textpm0.036}
    & 0.241{\tiny\textpm0.021}
    & 0.339{\tiny\textpm0.039}
    & 0.243{\tiny\textpm0.022}
    \\

    $\mathcal{S}_c$ & $\mathcal{R}$
    & 0.305{\tiny\textpm0.030}
    & 0.261{\tiny\textpm0.021}
    & 0.312{\tiny\textpm0.033}
    & 0.262{\tiny\textpm0.020}
    \\


    $\mathcal{S}_c$ & $\mathcal{S}_p, \mathcal{S}_w$
    & 0.310{\tiny\textpm0.060}
    & \underline{0.236{\tiny\textpm0.020}}
    & 0.321{\tiny\textpm0.064}
    & 0.239{\tiny\textpm0.022}
    \\

    $\mathcal{S}_c$ & $\mathcal{E}, \mathcal{R}$
    & \underline{0.289{\tiny\textpm0.031}}
    & 0.252{\tiny\textpm0.016}
    & \underline{0.294{\tiny\textpm0.034}}
    & 0.254{\tiny\textpm0.016}
    \\

    $\mathcal{S}_c$ & $\mathcal{S}_p, \mathcal{S}_w, \mathcal{E}$
    & 0.294{\tiny\textpm0.043}
    & \underline{0.236{\tiny\textpm0.017}}
    & 0.311{\tiny\textpm0.047}
    & \underline{0.238{\tiny\textpm0.018}}
    \\

    $\mathcal{S}_c$ & $\mathcal{S}_p, \mathcal{S}_w, \mathcal{R}$
    & 0.305{\tiny\textpm0.036}
    & 0.255{\tiny\textpm0.019}
    & 0.308{\tiny\textpm0.037}
    & 0.255{\tiny\textpm0.019}
    \\

    $\mathcal{S}_c$ & $\mathcal{S}_p, \mathcal{S}_w, \mathcal{E}, \mathcal{R}$
    & \textbf{0.282{\tiny\textpm0.032}}
    & 0.250{\tiny\textpm0.017}
    & \textbf{0.287{\tiny\textpm0.035}}
    & 0.250{\tiny\textpm0.017}
    \\


    \bottomrule
    
    \end{tabular}
    \begin{tablenotes}\scriptsize
    \item[*] This model requires two parallel Brownian bridge processes.    
    \end{tablenotes}
    \end{threeparttable}
    
    \label{tab:ablation}
\end{table}

\subsection{Image Variability} \label{sec:variability}

\begin{figure}[t]
    \centering
    \includegraphics[width=0.95\linewidth]{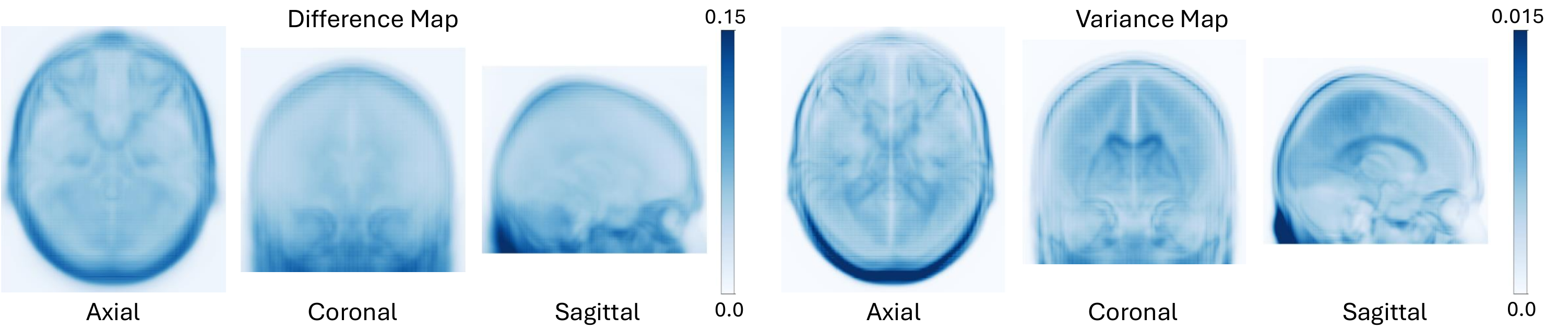}
    \caption{We show the mean absolute difference between synthetic and original MRIs, and voxel-wise variance across synthetic MRIs across five random seeds among the whole test set, averaged along three anatomical axes, with 
    darker colors indicating higher variability.
    }
    \label{fig:varibaility}
\end{figure}

While we aim to generate MRI scans that conform to the shape of given cortical surfaces, it is desirable for applications like dataset augmentation to have high diversity in the remaining image regions. We generated five synthetic MRIs per subject in the test set using different random seeds and calculated the average variance in individual voxels across the synthetic MRIs. Additionally, we computed the mean absolute difference between these synthetic scans and the corresponding real MRIs. As illustrated in \Cref{fig:varibaility}, the largest variation can be expected in the skull regions, whereas the cortical regions exhibit minimal differences to the original data and also comparably low variance. This indicates that Cor2Vox effectively generates realistic MRI scans that accurately align with the cortical surfaces while 
promoting diversity in the remaining image parts. 

\subsection{Synthesizing Cortical Atrophy}
\begin{figure}[t]
    \centering
    \includegraphics[width=\linewidth]{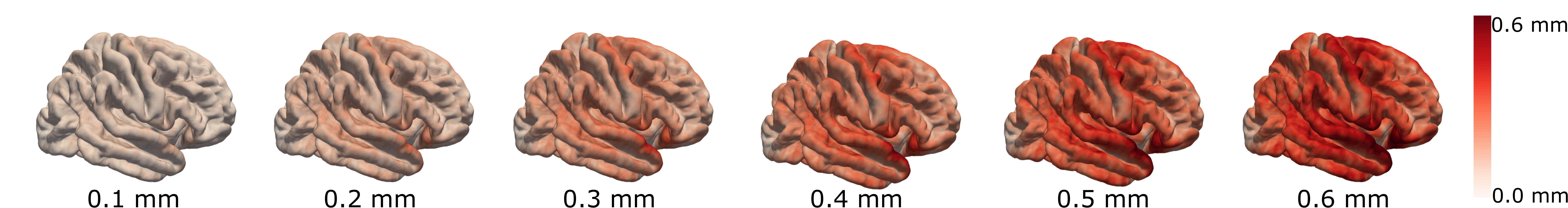}
    \caption{Synthetic global atrophy in the range of 0.1--0.6 mm recovered from MRIs generated by Cor2Vox. We show vertex-wise mean values based on 124 cognitively normal test cases. The closer the recovered values to the introduced atrophy, the better.}
    \label{fig:atrophy}
\end{figure}

The deliberate combination of white matter and pial surfaces in Cor2Vox enables the creation of synthetic datasets with simulated cortical thickness changes. Inspired by previous work on cortical atrophy simulation~\cite{Rusak2022benchmark}, we mimicked the process of global cortical thinning in the range of 0.1--0.6 mm by deforming pial surfaces from cognitively normal subjects in our test set ($n=124$) inside towards the white matter boundary. After generating matching MRIs from the altered surfaces (Cor2Vox) and reconstructing respective cortical surfaces (V2C-Flow), we compared the recovered change in cortical thickness (measured bi-directionally~\cite{fischl2000}) to the introduced atrophy, see \Cref{fig:atrophy}. Except for a small occipital area, the introduced changes in the subvoxel range were well recovered, supporting the applicability of Cor2Vox for algorithm benchmarking.

\section{Conclusion}
We introduced Cor2Vox, a novel method for 3D brain MRI generation based on cortical surfaces. For the first time, we leveraged a shape-to-image Brownian bridge diffusion process for synthesizing anatomically plausible 3D medical scans. We extended the Brownian diffusion process by incorporating multiple complementary shape conditions to enhance anatomical alignment.
By conditioning on realistic 3D shapes, 
Cor2Vox effectively addresses the problem of anatomical implausibilities commonly seen in synthetic brain MRIs. Our results demonstrated state-of-the-art image quality and significant improvements in geometric accuracy compared to existing methods. Moreover, we conducted a comprehensive ablation study and showcased the application of Cor2Vox for simulating individual cortical atrophy. Our findings highlight the critical role of realistic anatomical modeling in synthetic medical image generation and pave the way for advanced data augmentation and algorithm benchmarking. 


\begin{credits}
\subsubsection{\ackname} This research was supported by the German Research Foundation (DFG) and the Munich Center for Machine Learning (MCML). We gratefully acknowledge the computational resources provided by the Leibniz Supercomputing Centre (www.lrz.de).


\end{credits}
%
%

\bibliographystyle{splncs04}
\bibliography{bibliography}

\end{document}